\documentclass{article}
\usepackage{spconf,amsmath,graphicx}
\usepackage{gensymb}
\usepackage{booktabs}
\usepackage{url}
\usepackage{siunitx}
\usepackage{subcaption}
\usepackage{tablefootnote}

\def\U{{\mathbf U}}
\def\x{{\mathbf x}}
\def\y{{\mathbf y}}

\DeclareMathOperator*{\argmax}{arg\,max}

\title{AUDIO-VISUAL SPEECH RECOGNITION WITH A HYBRID CTC/ATTENTION ARCHITECTURE}
\name{\begin{tabular}{c}Stavros Petridis$^{1,3}$, Themos Stafylakis$^{2,4}$, Pingchuan Ma$^1$\\
Georgios Tzimiropoulos$^{2,3}$, Maja Pantic$^{1,3}$\end{tabular}}
\address{$^1$Dept. of Computing, Imperial College London, UK \\
$^2$Computer Vision Laboratory, University of Nottingham, UK \\
$^3$Samsung AI Center, Cambridge, UK \\
$^4$Omilia - Conversational Intelligence, Athens, Greece \\
stavros.petridis04@imperial.ac.uk, tstafylakis@omilia.com \\
}
\begin{document}
%\ninept
%
\maketitle
\begin{abstract}
Recent works in speech recognition rely either on connectionist temporal classification (CTC) or sequence-to-sequence models for character-level recognition. CTC assumes conditional independence of individual characters, whereas attention-based models can provide nonsequential alignments. Therefore, we could use a CTC loss in combination with an attention-based model in order to force monotonic alignments and at the same time get rid of the conditional independence assumption. In this paper, we use the recently proposed hybrid CTC/attention architecture for audio-visual recognition of speech in-the-wild. To the best of our knowledge, this is the first time that such a hybrid architecture architecture is used for audio-visual recognition of speech. We use the LRS2 database and show that the proposed audio-visual model leads to an 1.3\% absolute decrease in word error rate over the audio-only model and achieves the new state-of-the-art performance on LRS2 database (7\% word error rate). We also observe that the audio-visual model significantly outperforms the audio-based model (up to 32.9\% absolute improvement in word error rate)  for several different types of noise as the signal-to-noise ratio decreases.

\end{abstract}
\begin{keywords}
Audiovisual Speech Recognition, Attention Architectures, CTC, Audiovisual Fusion
\end{keywords}
\section{Introduction}
\label{sec:intro}

\begin{figure*}
    \centering
    \begin{subfigure}[b]{0.35\textwidth}
        \includegraphics[width=\textwidth]{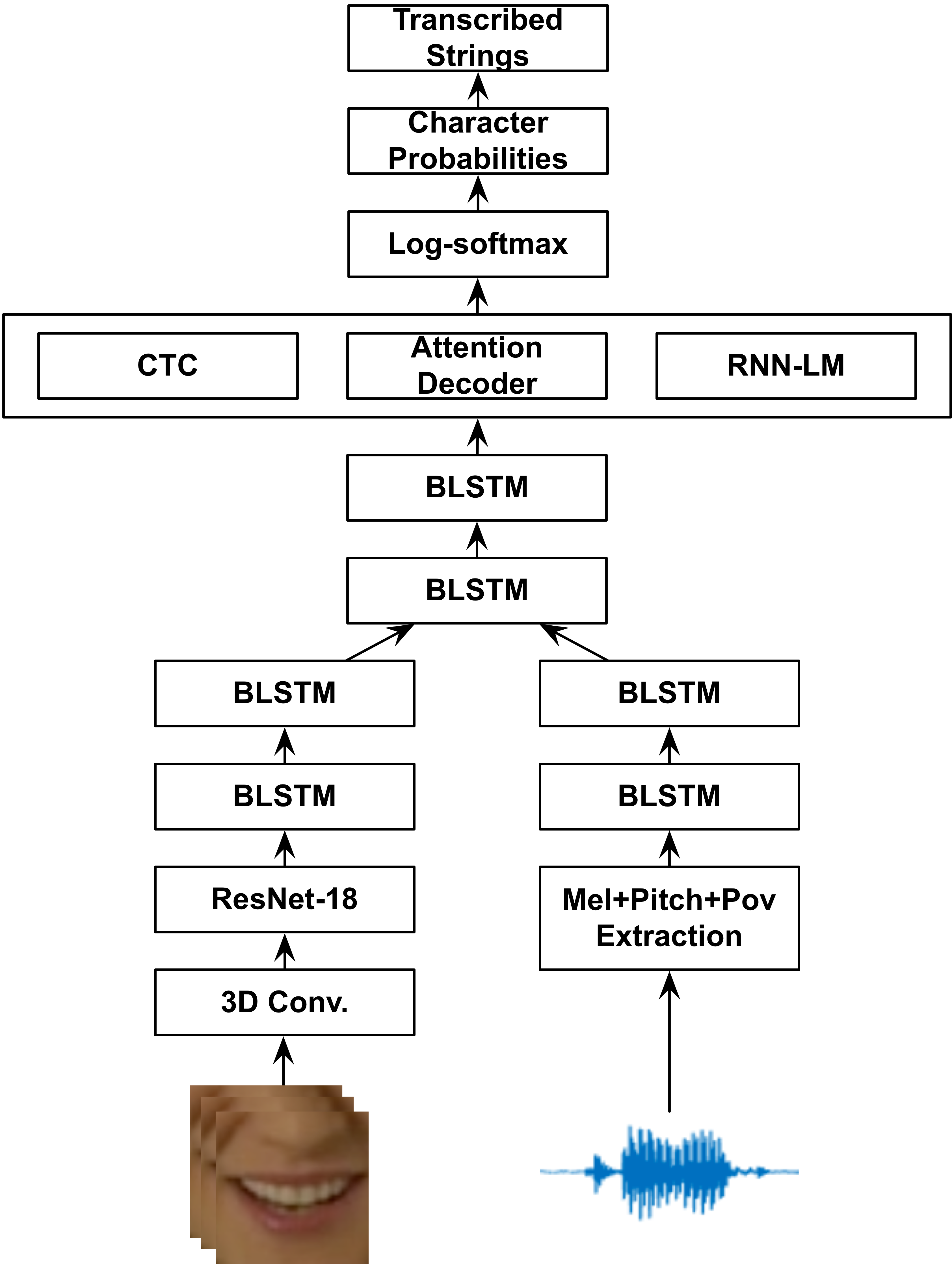}
        \caption{Early Fusion}
        \label{fig:earlyFusion}
    \end{subfigure}
    ~ %add desired spacing between images, e. g. ~, \quad, \qquad, \hfill etc. 
      %(or a blank line to force the subfigure onto a new line)
    \begin{subfigure}[b]{0.48\textwidth}
        \includegraphics[width=\textwidth]{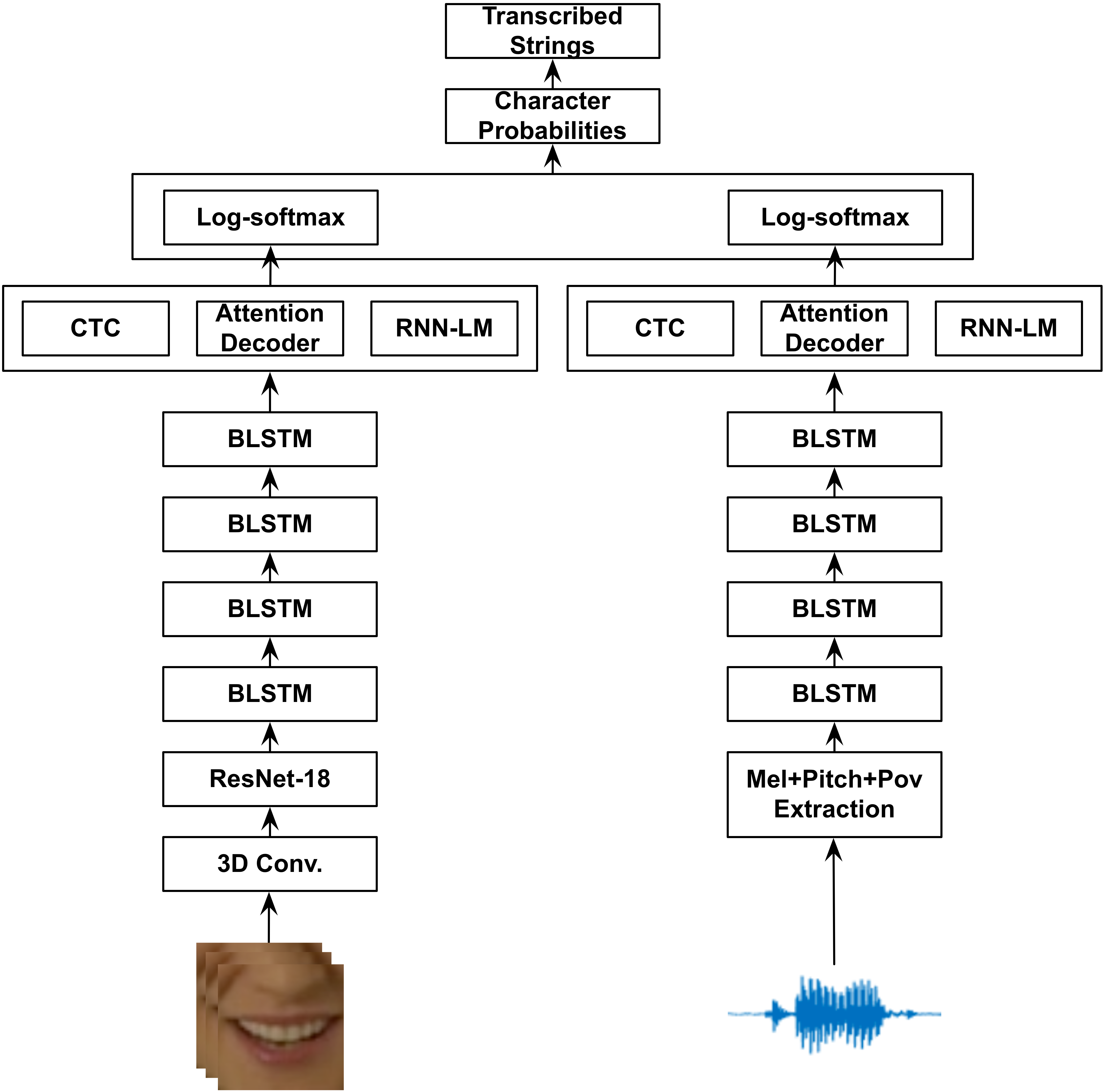}
        \caption{Late Fusion}
        \label{fig:lateFusion}
    \end{subfigure}
    ~ %add desired spacing between images, e. g. ~, \quad, \qquad, \hfill etc. 
    %(or a blank line to force the subfigure onto a new line)
    \caption{Architectures considered in this work. The encoder  consists of a stack of BLSTMs, whereas a joint CTC/attention approach is followed for decoding together with an external language model.}\label{fig:archit}
\end{figure*}

Traditional audiovisual fusion systems consist of two stages, feature extraction from the image and audio signals and combination of the features for joint classification \cite{Potamianos2003,Dupont2000,predBasedAVfusion}. Although decades of 
research in acoustic speech recognition have resulted in a standard set of audio features, there is not a standard set of visual features yet. This issue has been recently addressed by the introduction of deep learning in this field. In the first generation of deep models, deep bottleneck architectures \cite{ngiam2011multimodal,hu2016temporal,ninomiya2015integration,mroueh2015,takashima2016audio,petridisPantic_icassp2016} were used to reduce the dimensionality of various visual and audio features extracted from the mouth regions of interest (ROI) and the audio signal. Then these features are fed to a classifier like a support vector machine or a Hidden Markov Model. 

Recently, few deep models have been presented which extract features directly from the mouth ROI pixels. The main approaches followed can be divided into two groups. In the first one, fully connected layers are used to extract features and LSTM layers model the temporal dynamics of the sequence \cite{petridis2017deepVisualSpeech,wand2016lipreading}. In the second group, a 3D convolutional layer is used followed either by standard convolutional layers \cite{assael2016lipnet,shillingford2018large} or residual networks (ResNet) \cite{stafylakis2017combining} combined with LSTMs or GRUs. 

These works have also been extended to audio-visual models. Chung et al. \cite{Chung17cvpr} applied an attention mechanism to both the mouth ROIs and MFCCs for continuous speech recognition. Petridis et al. \cite{end2endAV} used fully connected layers together with LSTMs are used in order to extract features directly from raw images and spectrograms and perform classification on the OuluVS2 database \cite{Anina2015}. This method has been extended to extract features directly from raw images and audio waveforms using ResNets and bidirectional gated recurrent units (BGRUs) \cite{petridis2018end} and achieves the state-of-the-art performance on the LRW dataset \cite{chung2016lip} for isolated within context word recognition in-the-wild.

In this work, we use ResNets to extract features directly from the mouth ROIs together with a hybrid CTC/attention architecture \cite{watanabe2017hybrid} for audio-visual continuous speech recognition in-the-wild.  Attention-based  speech recognition  uses  an  attention  mechanism  to  find  an  alignment between each element of the output sequence and the hidden states  generated  by  the    encoder  network  for  each frame  of  acoustic/visual  input. The main problem with this approach is that it allows non-sequential alignments. This can be addressed using a connectionist temporal classification (CTC) objective (which allows for a strictly monotonic alignment) together with the attention-based encoder-decoder. This hybrid CTC/attention architecture has been successfully used in acoustic speech recognition \cite{watanabe2017hybrid}. A similar idea has been explored in \cite{LCAnet} where a cascaded CTC-attention model is proposed for visual speech recognition on the GRID database, which has been recorded in a lab environment. To the best of our knowledge, this is the first work which uses a hybrid CTC/attention architecture for audio-visual speech recognition in-the-wild. For this purpose, we use the LRS2 database, which is the largest publicly available database of continuous audio-visual speech in-the-wild.

\begin{table}[t]
\renewcommand{\arraystretch}{1.1}
\renewcommand{\tabcolsep}{7pt}
\caption{ Statistics of the LRS2 dataset.   }
\label{tab:LRS2stats}
\centering
\begin{tabular}{lccc}
\toprule  Set & No. & No.& Vocabulary   \\
& Utterances  & Words & \\

\midrule Pre-training  &  96318	&2064118	&41427 \\
Training  & 45839	&329180	&17660\\
Validation  & 1082	&7866	&1984\\
Test  & 1243 &	6663	&1698  \\

\bottomrule

\end{tabular} 

\end{table}

The proposed system, Fig. \ref{fig:archit}, results in an absolute decrease of 6.9\% in 
word error rate (WER) for visual-only speech recognition over the state-of-the-art on LRS2 (without using external datasets). The audio-visual model leads to a 1.3\% absolute improvement over the audio-only model in clean audio conditions and achieves the new state-of-the-art audio-visual performance (7\% WER) outperforming even models which were pre-trained on external datasets. We also investigate the effect of different types of noise at varying levels of signal-to-noise ratio (SNR), from -5dB to 20dB, on the audio-only and and audio-visual models. As expected the audio-visual model is more robust to all types of noise leading to an absolute decrease in WER of up to 32.9\% at high SNR levels over the audio-based model.

\begin{figure*}[htp]

\centering
\includegraphics[width=.13\textwidth]{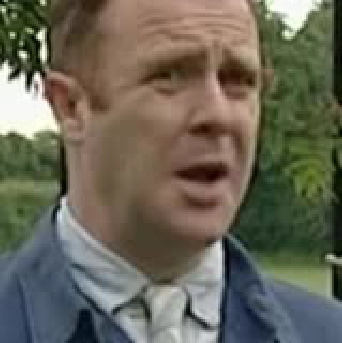}\hfill
\includegraphics[width=.13\textwidth]{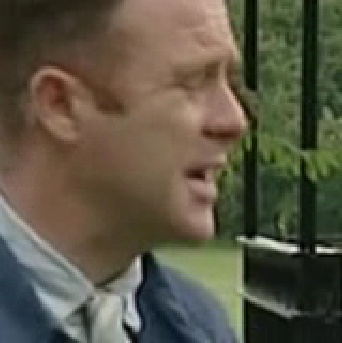}\hfill
\includegraphics[width=.13\textwidth]{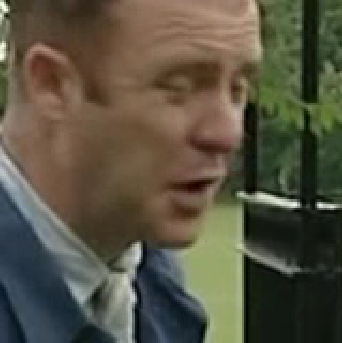}\hfill
\includegraphics[width=.13\textwidth]{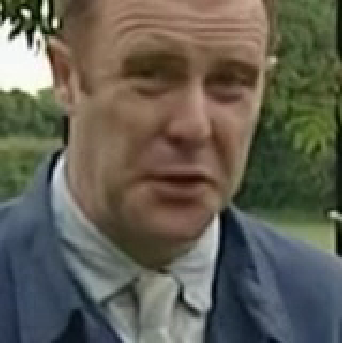}\hfill
\includegraphics[width=.13\textwidth]{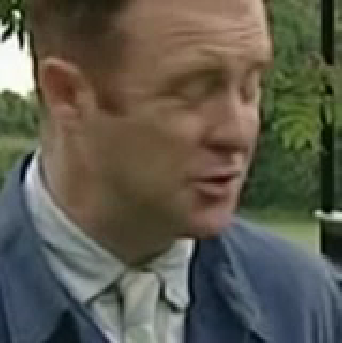}\hfill
\includegraphics[width=.13\textwidth]{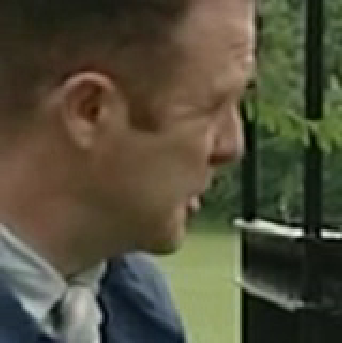}
\caption{Example of significant head pose variation from the LRS2 dataset.}
\label{fig:exHeadMovement}

\end{figure*}

% \begin{figure*}[htp]

% \centering
% \includegraphics[width=.13\textwidth]{Figures/5540444676275685005_00014_fr_1.png}\hfill
% \includegraphics[width=.13\textwidth]{Figures/5540444676275685005_00014_fr_12.png}\hfill
% \includegraphics[width=.13\textwidth]{Figures/5540444676275685005_00014_fr_48.png}\hfill
% \includegraphics[width=.13\textwidth]{Figures/5540444676275685005_00014_fr_58.png}\hfill
% \includegraphics[width=.13\textwidth]{Figures/5540444676275685005_00014_fr_72.png}\hfill
% \includegraphics[width=.13\textwidth]{Figures/5540444676275685005_00014_fr_120.png}
% \caption{Example of significant head pose variation from the LRS2 dataset.}
% \label{fig:exHeadMovement}

% \end{figure*}

% \begin{figure*}[htp]

% \centering
% \includegraphics[width=.13\textwidth]{Figures/5581310431103671083_00003_fr_1.png}\hfill
% \includegraphics[width=.13\textwidth]{Figures/5581310431103671083_00003_fr_13.png}\hfill
% \includegraphics[width=.13\textwidth]{Figures/5581310431103671083_00003_fr_25.png}\hfill
% \includegraphics[width=.13\textwidth]{Figures/5581310431103671083_00003_fr_37.png}\hfill
% \includegraphics[width=.13\textwidth]{Figures/5581310431103671083_00003_fr_49.png}\hfill
% \includegraphics[width=.13\textwidth]{Figures/5581310431103671083_00003_fr_58.png}
% \caption{Example of profile view from the LRS2 dataset.}
% \label{fig:exProfileView}

% \end{figure*}

\section{LRS2 DATABASE}
\label{sec:LRS2}

For the purposes of this study we use the Lip Reading Sentences 2 (LRS2) database \cite{Chung17cvpr,Chung17bmvc} which is the largest publicly available dataset for lip reading sentences in-the-wild. The database consists of short segments (up to 6.2 seconds) from BBC programmes, mainly news and talk shows. It is a very challenging set since it contains thousands of speakers and large variation in head pose (from frontal to profile) and illumination. 

The dataset contains more than 2 million words and more than 140K utterances. An example of large head pose variation can be seen in Fig. \ref{fig:exHeadMovement}. The dataset is already divided into training, validation and test sets and also contains a pre-training set which contains longer segments (up to 181.8 seconds) which can be used to pre-train a model. Details about the dataset can be found in Table \ref{tab:LRS2stats}.

\section{Architecture}

\subsection{Features}

\textbf{Visual Features:}
The visual feature extractor is based on the model proposed in \cite{stafylakis2017combining}.
It consists of a spatiotemporal convolution with a filter width of 5 frames, which is capable
of capturing the short-term dynamics of the mouth region, followed by an 18-layer residual network (ResNet).
The ResNet drops progressively the spatial dimensionality until its output becomes a single dimensional tensor per time step. 
The output of the last fully connected of the ResNet is used as the visual feature representation. The features are extracted at 25 frames per second (fps) which is the frame rate of the input video.

\textbf{Audio Features:} We use 80 log Mel features together with pitch, delta pitch and probability of voicing, so there are 83 features in total. The features are extracted using a 25ms Hamming window with stride 10ms which results in 100 fps.

\subsection{Hybrid CTC/Attention}
\label{sec:typestyle}
To map a set of input sequences such as audio or video streams to corresponding output sequences, we consider a hybrid CTC/attention architecture \cite{watanabe2017hybrid} in this paper. This architecture uses a typical encoder-decoder attention structure. A stack of Bidirectional Long Short Term Memory Networks (BLSTMs)   is employed in the encoder to convert input streams $\x=(x_1, ..., x_T)$ into frame-wise hidden feature representations. These features are then consumed by a joint decoder including a recurrent neural network language model (RNN-LM), attention and CTC mechanisms to output a label sequence $\y=(y_1, ..., y_L)$. To perform alignment between input frames and output characters, we use a location-based attention mechanism, which takes into account both content and location information for selecting the next step in the input sequence \cite{chorowski2015attention}.

This architecture is proven to be advantageous for three reasons. First, the attention mechanism is built without any conditional independence assumptions. This helps build a more precise model. Second, a new blank token introduced in CTC is capable of directly transcribing between variable sequences without any intermediate annotation. Furthermore, the joint architecture introduces CTC for satisfying the monotonic alignment property required in speech recognition. 

The joint architecture shares the same encoder but uses separate mechanisms in the decoder, which can be considered as multi-task learning. During training, the objective function is performed by a linear combination of the CTC and attention objectives, which is computed as follows:
\begin{align}
\mathcal{L} = \alpha logp_{ctc}(\y|\x) + (1 - \alpha)logp_{att}(\y|\x)
\label{eq:trainingCTCweight}
\end{align}
where $\alpha$  controls the relative weight in CTC and attention mechanisms.

In the decoding phase, a joint CTC/attention approach is employed. This approach overcomes the drawback of the attention-only approach that has non-monotonic alignment and end-of-sentence detection issues. We obtain a joint score based on attention probabilities and CTC probabilities for decoding character-level sequences. The most probable output hypothesis $\hat{y}$ is computed as follows:
\begin{align}
\hat{\y}=\argmax_{\y\in \U}\{\lambda logp_{ctc}(\y|\x)+(1-\lambda)logp_{att}(\y|\x)\} \label{eq:outputHypothesis}
\end{align} 
where $\lambda$\footnote{We follow the notation of the ESPnet toolkit\cite{espnet} where the relative weight of CTC during training can be different than the CTC weight during decoding.} is the weight of CTC and $\U$ is the set of labels plus an extra end-of-sentence label. This approach also includes a beam search algorithm that recursively advances to the next label using the joint score of each partial hypothesis. 

For decoding, we include a character-level RNN-LM, which we train on LRS2 (train and pretrain sets) as well as on LibriSpeech \cite{panayotov2015librispeech}. The RNN-LM is incorporated through shallow fusion \cite{kannan2017analysis}, which is described as follows:
\begin{align}
logp^{hyb}(\y|\x)=&\lambda logp_{ctc}(\y|\x)+(1-\lambda)logp_{att}(\y|\x) \nonumber \\
&+\beta logp_{RNN-LM}(\y) \label{eq:combinedProb} 
\end{align}
\begin{align}
\hat{\y}^\star=&\argmax_{\y\in \U} \{logp^{hyb}(\y|\x)\}
\end{align}
where $\beta$ is a relative weight for the RNN-LM model.

\subsection{Fusion Types}
Two types of fusion are considered in this work, early fusion and late fusion as shown in Fig. \ref{fig:archit}. In early fusion, audio and visual features 
are concatenated inside the encoder as shown in Fig. \ref{fig:earlyFusion}. They are fed to two independent 2-layer BLSTMs whose outputs are concatenated. This is followed by another 2-layer BLTSM which produces the hidden representations fed to the CTC/attention decoder. 

In late fusion, Fig. \ref{fig:lateFusion}, audio and video are modeled independently by separate encoder-decoder architectures and then the generated character probabilities are fused as follows: 

\begin{align}
logp^{hyb}_{late\_fusion} = \gamma logp^{hyb}_{audio}+ (1-\gamma)logp^{hyb}_{visual}
\label{eq:latefusion}
\end{align} 
where $\gamma$, from 0.0 to 1.0, is a hyper-parameter to control the relative weight between audio and visual probabilities. 

\section{EXPERIMENTAL SETUP}

\subsection{Pre-processing}
The first step is the extraction of the mouth ROI from the LRS2 dataset. Since the mouth ROIs are already centered, a fixed bounding box of 130 by 80 is used for all videos, which is then resized to 122 by 122 (the input frame size of the ResNet is 112 by 112, using random cropping in training and the central patch in testing). Finally, the frames are transformed to grayscale and are normalized with respect to the overall mean and variance. The audio features are normalised by removing the mean and dividing by the standard deviation in each utterance.

\subsection{Evaluation Protocol}
Details about the data, which are already divided into training, validation and test sets, can be found in Table \ref{tab:LRS2stats}.
The utterances in the pre-training set correspond to part-sentences as well as multiple sentences, whereas the training set only consists of single full sentences. 

\section{Training}
Training is divided into 3 phases: first the visual
feature extractor is pre-trained on LRW 
and fine-tuned on LRS2. Then, the hybrid CTC/Attention model is trained with
the extracted visual and audio features. The ESPnet toolkit \cite{espnet} is used for training the hybrid CTC/attention architecture. Finally, an external language model
is trained using 2 text corpora.

\subsection{Pre-training of Visual Feature Extractor}

The ResNet is first pretrained on LRW for isolated word recognition.
A 2-layer BLSTM is added on top of the ResNet and the model is trained 
end-to-end (using a softmax output layer) as described in \cite{stafylakis2017combining}.
The Adam training algorithm \cite{kingma2014adam} is
used for end-to-end training with a mini-batch size of 36 sequences
and an initial learning rate of 0.0003. Early stopping
with a delay of 5 epochs is also used. Data augmentation
is also performed on the video sequences of mouth ROIs.
This is done by applying random cropping and horizontal
flips with probability 50\% to all frames of a given clip. 

The model is then further fine-tuned on the pretrain set LRS2. The pretrain set is useful for this purpose, not merely due to its large number of utterances, but also due to its more detailed annotation files, containing information about the (estimated) time each word begins and ends. Word boundaries permit us to excerpt fixed-duration video segments containing specific words and essentially mimic the LRW set-up. To this end, we select the 2000 most frequently appearing words containing at least 4 phonemes and we extract frame sequences of 1.5sec duration, having the target word in the center.

\subsection{Hybrid CTC/Attention}
The hybrid CTC/Attention model is trained for 20 epochs using Adadelta with  a mini-batch size of 10. Data augmentation is applied to the raw audio sequences before computing the mel and pitch features. During
training babble noise at different SNR levels (0 dB, 5 dB and 10
dB) from the NOISEX database \cite{varga1993assessment} might be added to the original audio clip. The selection
of one of the noise levels or the use of the clean audio is done
using a uniform distribution.

We also used label smoothing during training for the audio and visual models. There was no improvement on the validation set in case of audio-visual models so label smoothing was not applied in this case.

\subsection{Language Model}
The language model is trained by combining two different text corpora. The first one contains the transcriptions of the LibriSpeech corpus which contains 9.4 million words. The second one contains the transcriptions of the LRS2 pre-train set which contains more than 2 million words. 

\subsection{Parameters}
The default parameters of the ESPnet toolkit \cite{espnet} have been used. The only exception is the CTC weight $\alpha$ and $\lambda$ from eq. \ref{eq:trainingCTCweight} and \ref{eq:outputHypothesis}, respectively, which are optimised on the validation set. The optimal values for $\alpha$ and $\lambda$ are 0.2 and 0.1, respectively. The late fusion weight $\gamma$ from eq. \ref{eq:latefusion} is also optimised on the validation set and the optimal value found is 0.85. The language model weight $\beta$ is set to 0.4 for the audio and audio-visual models and 0.1 for the visual models. Finally, the width of beam search is set to 20.

\section{RESULTS}

\begin{table}[t]
\renewcommand{\arraystretch}{1.1}
\renewcommand{\tabcolsep}{7pt}
\caption{ Character error rate (CER) and Word Error Rate (WER) of the  Audio-only  (A), Video-only  (V) and Audio-Visual models (A + V) on the LRS2 database. * The model in \cite{afouras2018deep} is first pre-trained on a non-publicly available dataset.  }
\label{tab:resultsBBC}
\centering
\begin{tabular}{lcc}
\toprule  Stream & CER &  WER  \\

\midrule A   & 4.4 & 8.3 \\
A \cite{afouras2018deep}* & - & 9.7 \\
A \cite{sterpu2018attention} & 14.3 & 29.9 \\

\midrule 
V  & 42.1 & 63.5\\
V \cite{Chung17cvpr} \tablefootnote{Video-only results from \cite{Chung17cvpr} are reported on \url{http://www.robots.ox.ac.uk/~vgg/data/lip_reading/lrs2.html}.} & - & 70.4 \\
V \cite{afouras2018deep}* & - & 50.0 \\
\midrule
A + V (Late Fusion)  & 4.7 & 8.5 \\
A + V (Early Fusion)  & 3.6 & 7.0  \\
A + V \cite{afouras2018deep}* & -& 8.2 \\
A + V \cite{sterpu2018attention} &14.1 & 30.5 \\
\bottomrule

\end{tabular} 

\end{table}

\begin{figure*}
    \centering
    \begin{subfigure}[b]{0.35\textwidth}
        \includegraphics[width=\textwidth]{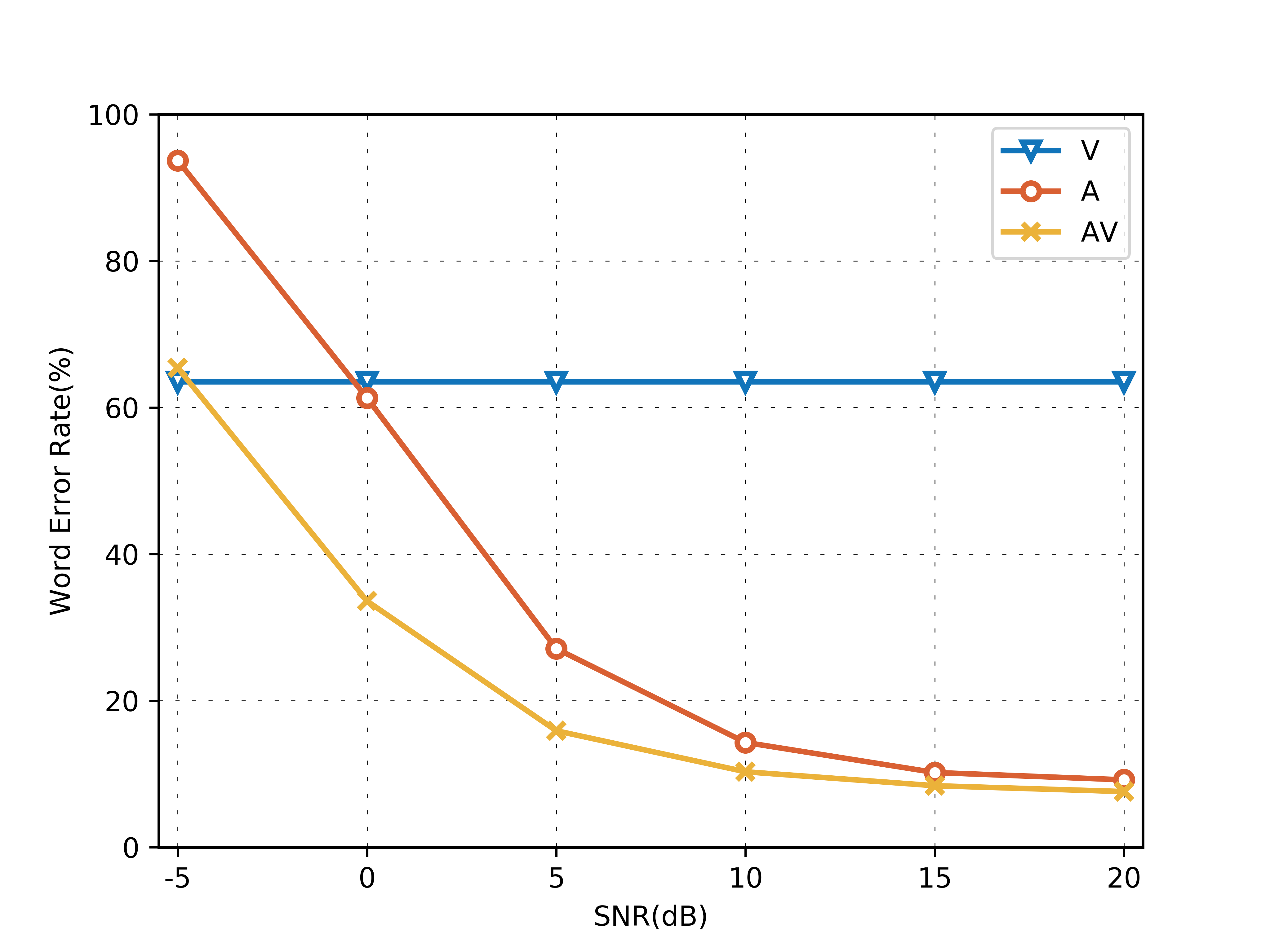}
        \caption{Cafe}
        \label{fig:cafe}
    \end{subfigure}
    ~ %add desired spacing between images, e. g. ~, \quad, \qquad, \hfill etc. 
      %(or a blank line to force the subfigure onto a new line)
    \begin{subfigure}[b]{0.35\textwidth}
        \includegraphics[width=\textwidth]{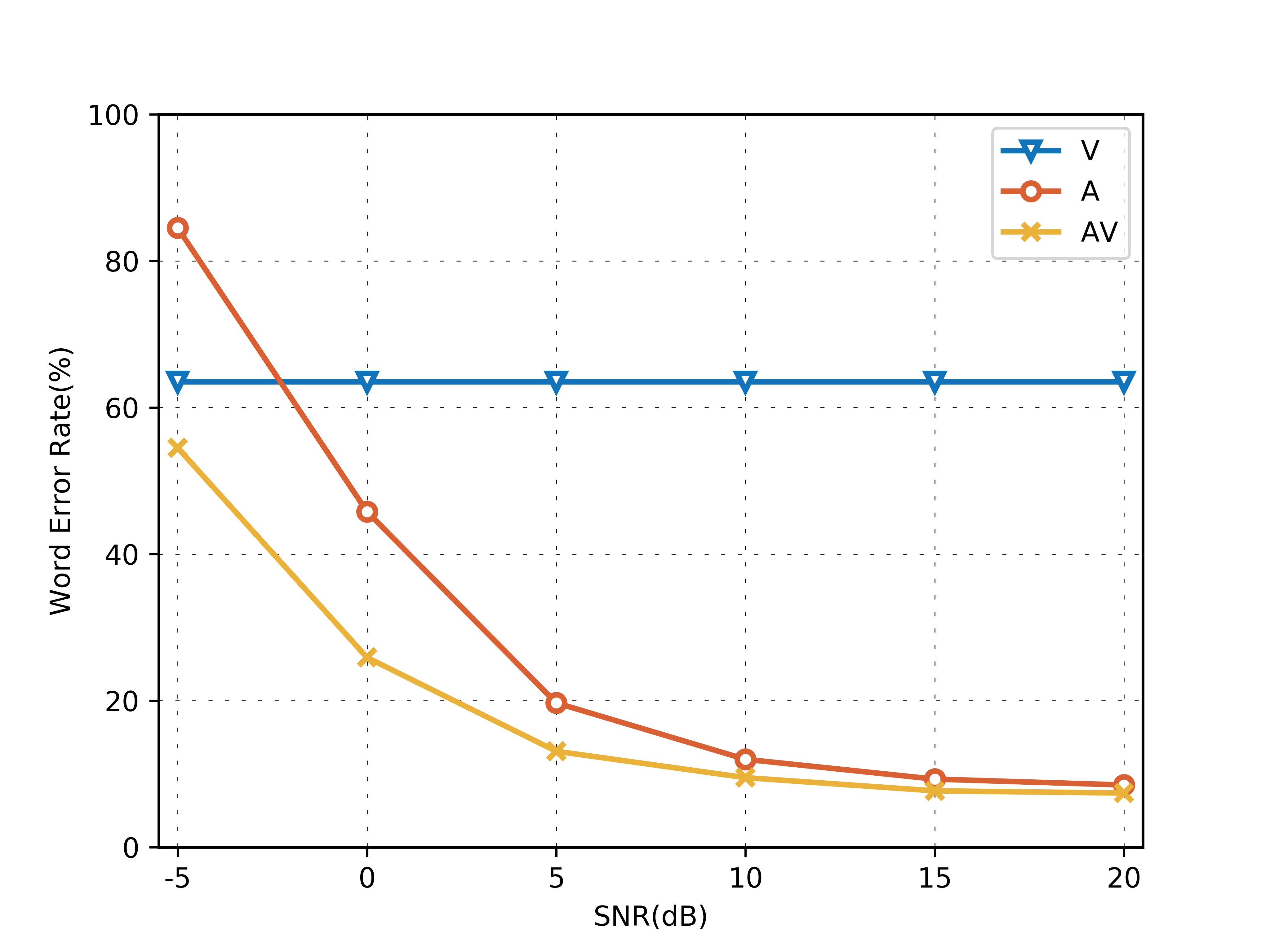}
        \caption{Street}
        \label{fig:street}
    \end{subfigure}
    ~ %add desired spacing between images, e. g. ~, \quad, \qquad, \hfill etc. 
    %(or a blank line to force the subfigure onto a new line)
    \begin{subfigure}[b]{0.35\textwidth}
        \includegraphics[width=\textwidth]{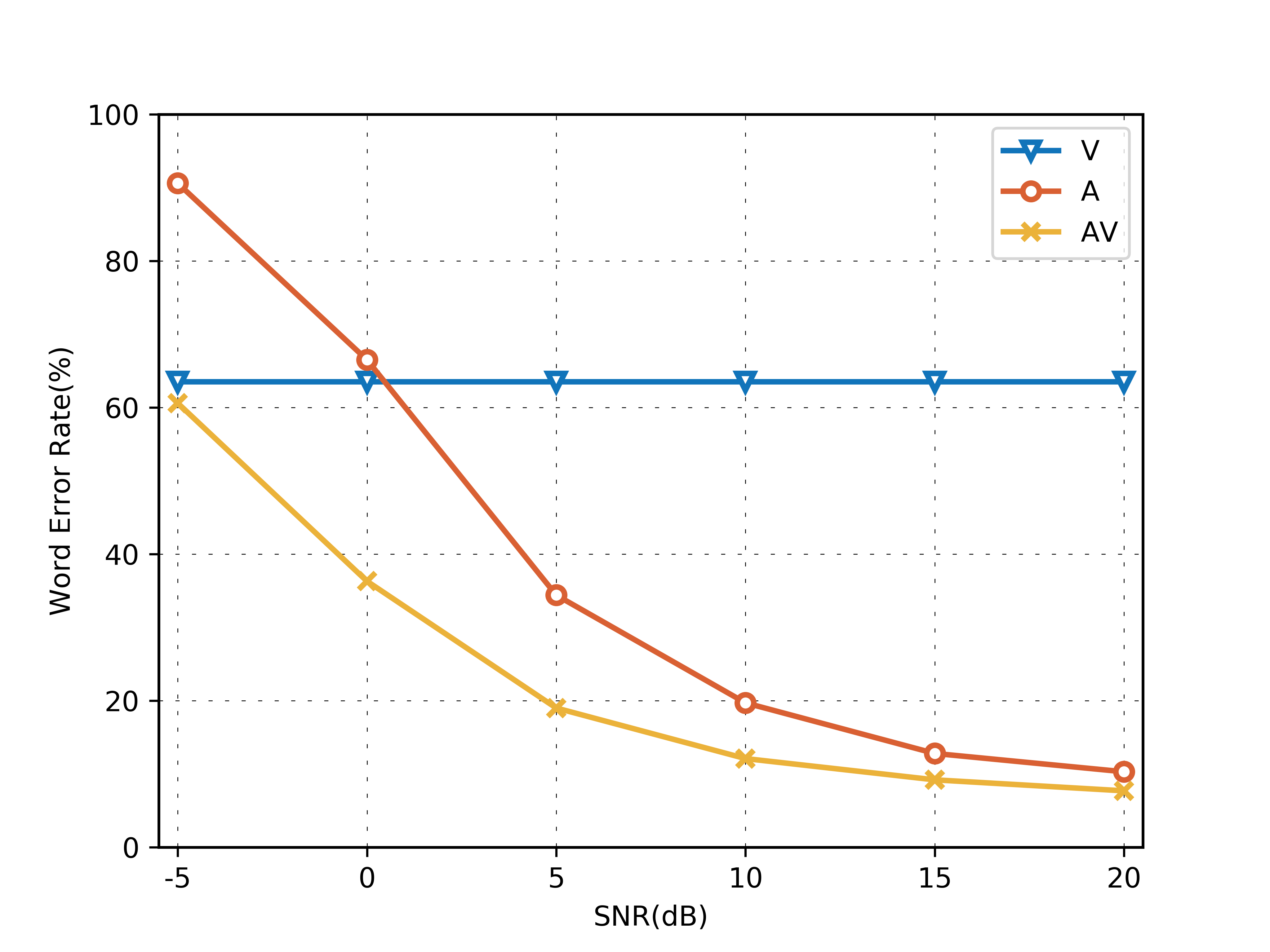}
        \caption{Pink}
        \label{fig:pink}
    \end{subfigure}
    \begin{subfigure}[b]{0.35\textwidth}
        \includegraphics[width=\textwidth]{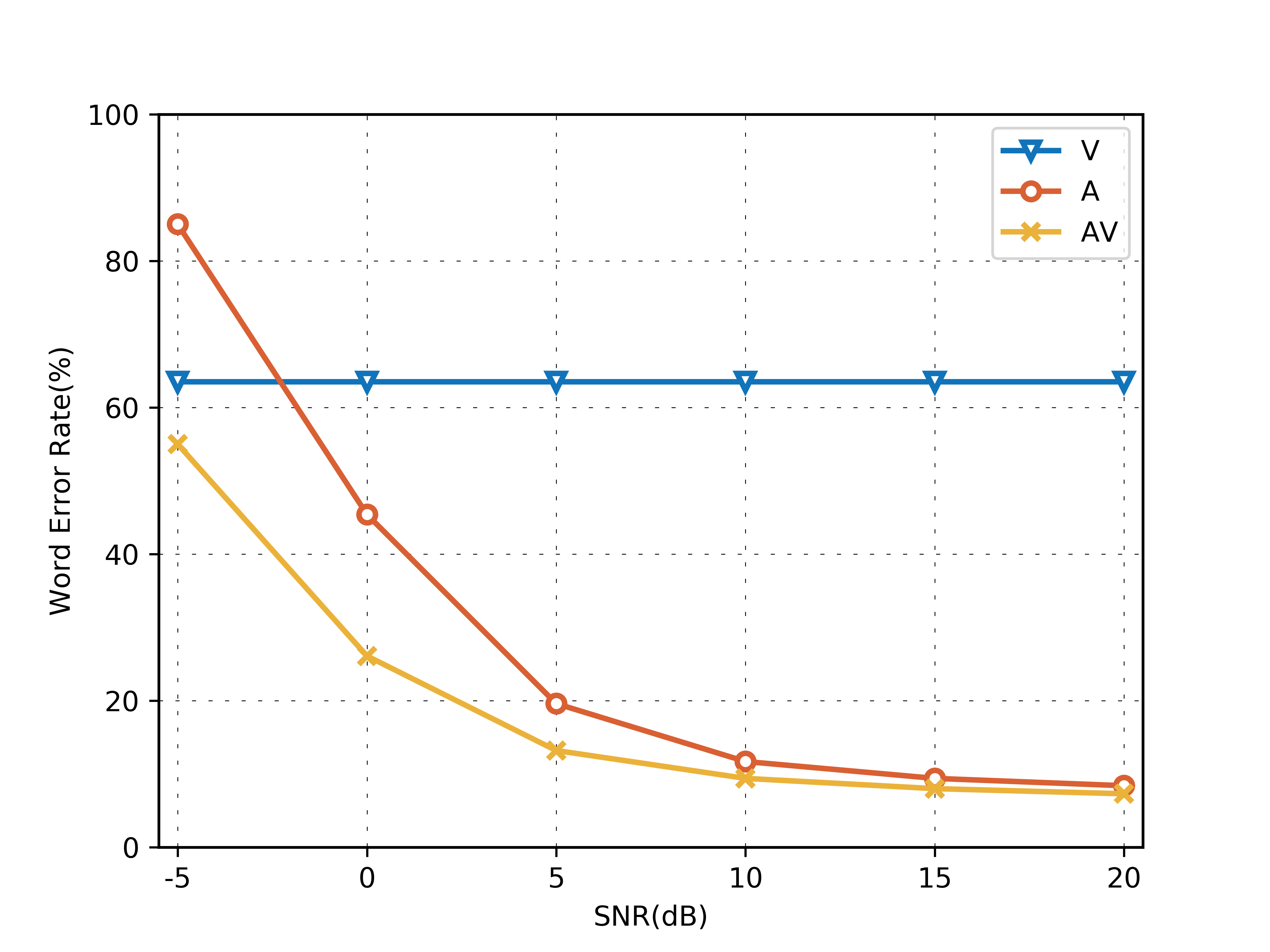}
        \caption{White}
        \label{fig:white}
    \end{subfigure}
    \begin{subfigure}[b]{0.35\textwidth}
        \includegraphics[width=\textwidth]{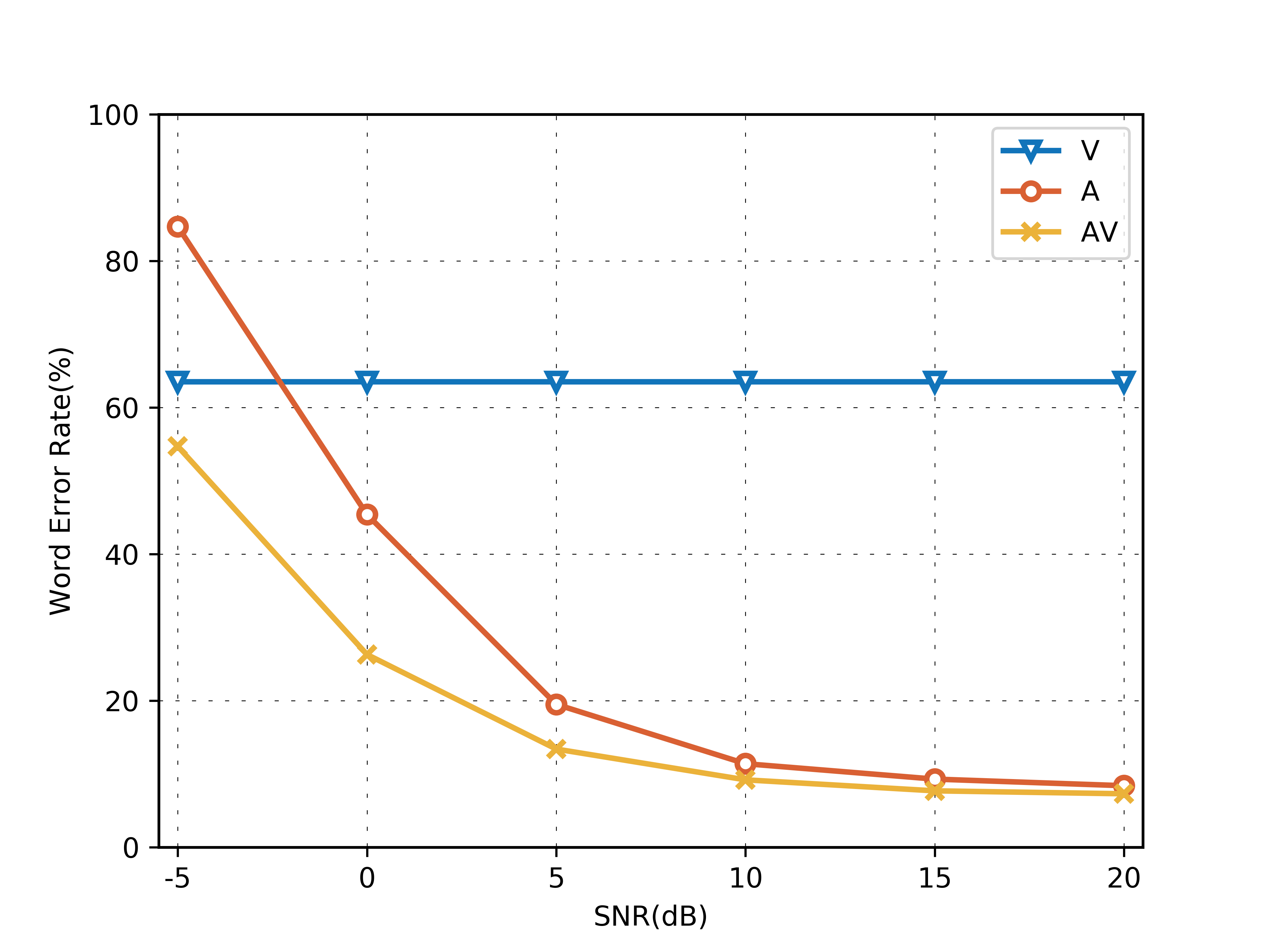}
        \caption{Doing Dishes}
        \label{fig:doingDishes}
    \end{subfigure}
    \begin{subfigure}[b]{0.35\textwidth}
        \includegraphics[width=\textwidth]{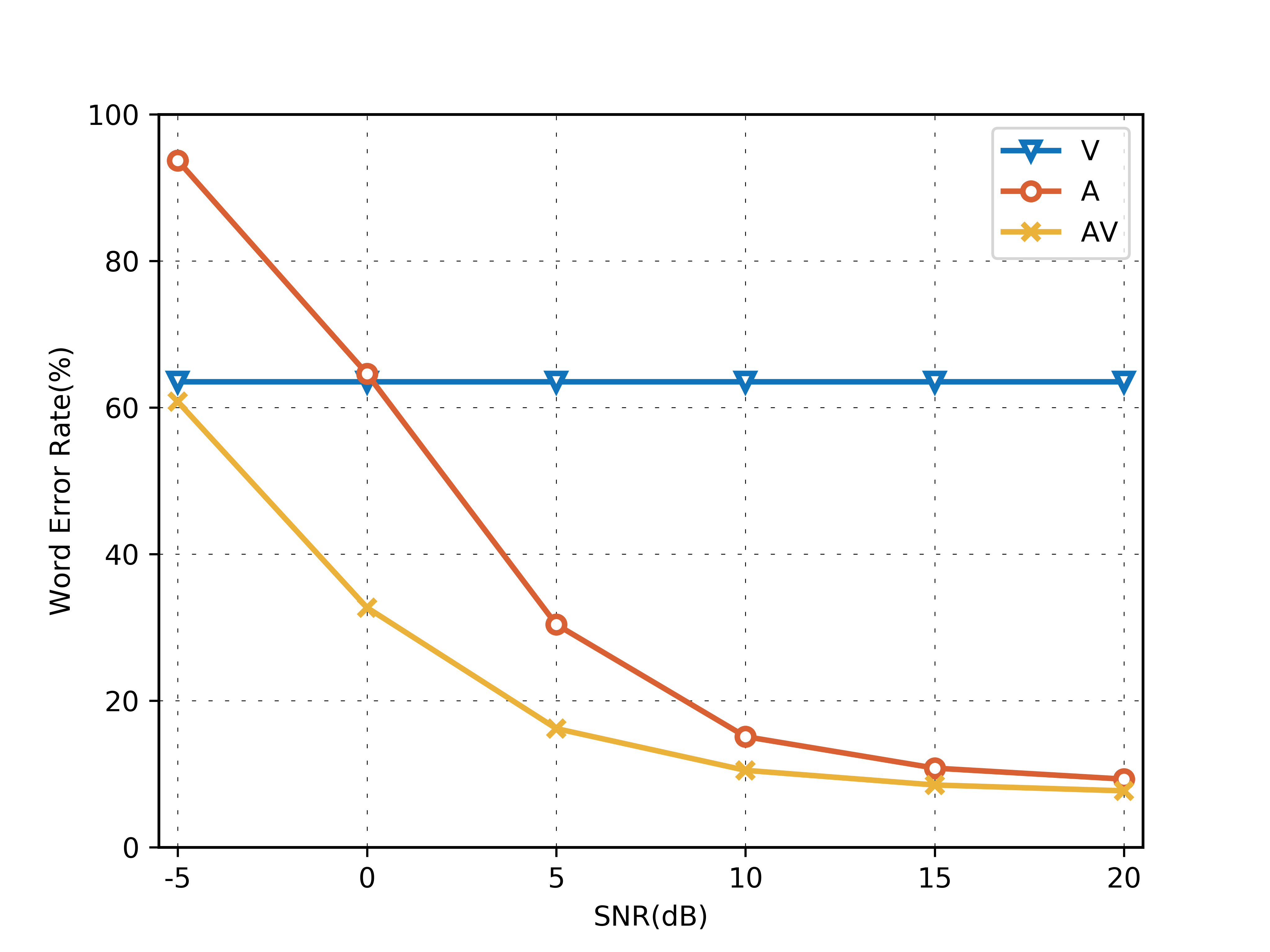}
        \caption{Construction Drilling}
        \label{fig:construction_drilling}
    \end{subfigure}
        \begin{subfigure}[b]{0.35\textwidth}
        \includegraphics[width=\textwidth]{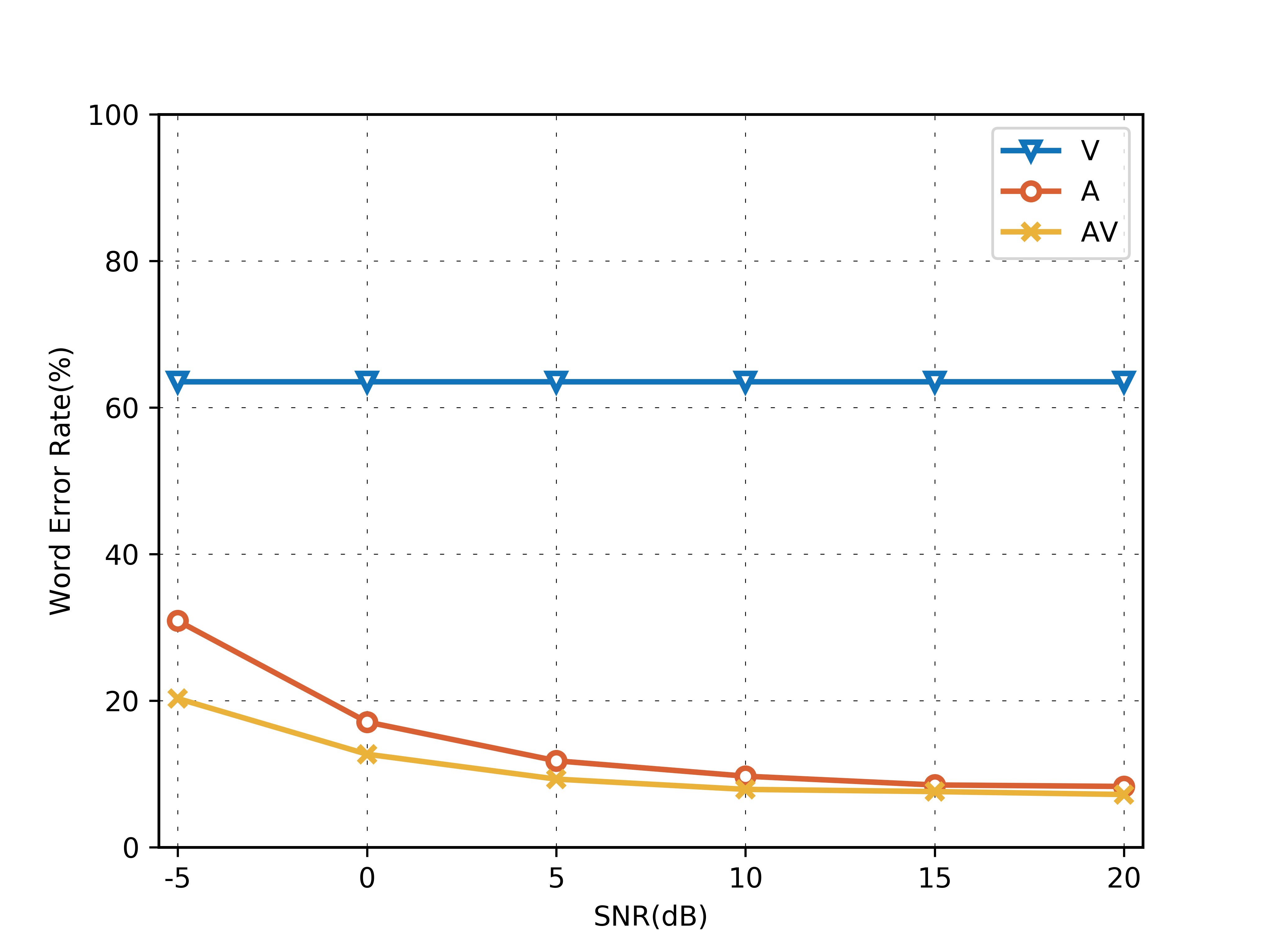}
        \caption{Car}
        \label{fig:carNoise}
    \end{subfigure}
    \begin{subfigure}[b]{0.35\textwidth}
        \includegraphics[width=\textwidth]{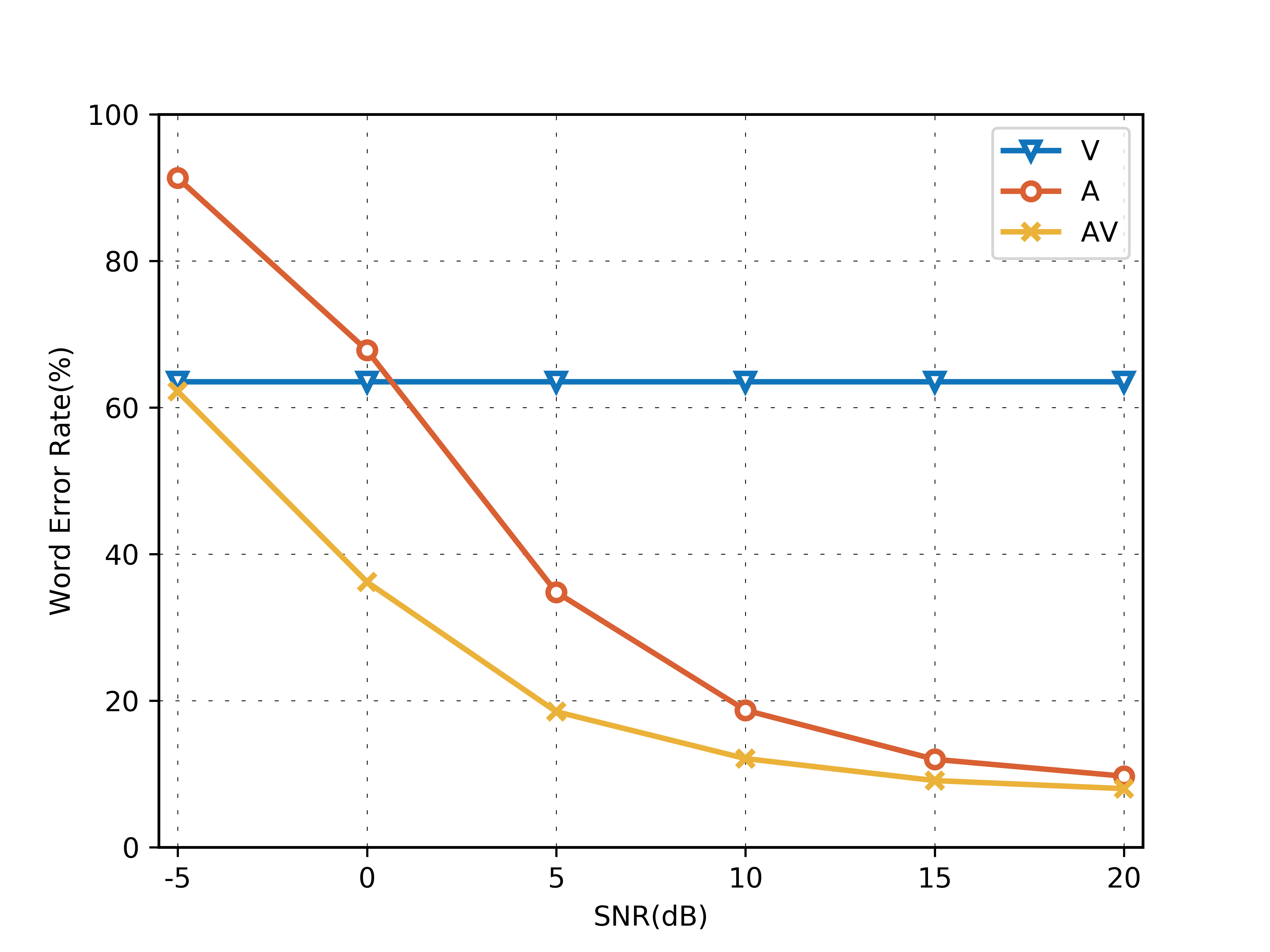}
        \caption{Train}
        \label{fig:train}
    \end{subfigure}
    \caption{WER of the video-only (V), audio-only (A) and audio-visual (AV) models as a function of the SNR for various noise types.}\label{fig:noiseTypes}
\end{figure*}

Results are shown in Table \ref{tab:resultsBBC}. We report the performance of the audio-only, visual-only and audiovisual models for both fusion types. It should be noted that the results shown correspond to visual features upsampled to 50 fps using linear interpolation. This is due to better performance observed on the validation set. Further upsampling did not improve the performance. The audio frame rate did not affect the performance so we report audio results at 100 fps. However, for all types of fusion we downsample audio to 50fps.

The proposed visual-only system results in an absolute improvement of 6.9\% in WER compared to \cite{Chung17cvpr} which is the state-of-the-art performance when training only on LRS2, i.e., without using any external databases. Afouras et al. \cite{afouras2018deep} achieve a much lower WER but their model is pre-trained on a non-publicly available dataset.

The audio-only model achieves an 8.3\% WER and 4.4\% CER. The audio-visual system using early fusion leads to an improvement over the audio-only models of 1.3\% and 0.8\% in WER and CER, respectively. Late fusion performs worse than early fusion resulting in an 8.5\% WER, possibly because it cannot directly model the correlation between audio and visual features. It is worth pointing out that both the audio-only and audio-visual models, which are trained only on LRS2, outperform \cite{afouras2018deep} which has been pre-trained on external databases. The WER of 7\% achieved by the audio-visual model is also the new state-of-the-art performance on LRS2.

In order to investigate the robustness to audio noise of the
audiovisual fusion approach we run experiments under
varying noise levels (using early fusion). The audio signal for each sequence
is corrupted by additive  noise so as the SNR varies from -5 dB to 20 dB. Five different noise types from \cite{loizou2007speech} are used, cafe, street, construction drilling, train and car noises. Three more noise types are used from \cite{speechcommands}, white, pink and doing dishes noises.

Results for the audio, visual and audiovisual models under noisy conditions are shown in Fig.  \ref{fig:noiseTypes}. The
video-only classifier (blue line)
is not affected by the addition of the audio noise and
therefore its performance remains constant over all noise
levels. On the other hand, as expected, the performance
of the audio-only model (red  line) is significantly affected. 
The WER of the audio-only model for all noise types lies between 8.3\% and 10.3\% at 20dB. On the other hand, the WER lies between 84.5\% and 93.7\% at -5dB. The only exception is the case of car noise, which corresponds to noise recorded inside a car driving at 60 miles per hour. The WER of the audio-only model for this type of noise is 30.9\%.

\looseness - 1
The audiovisual model (yellow line) 
is more robust to audio noise than the audio-only models. It results in an absolute improvement of up to 7.6\% (pink noise) under low noise levels (10 dB to 20 dB) but it significantly outperforms the audio-only model under high noise levels (-5 dB to 5 dB). In particular, it leads to an absolute improvement between 10.6\% (car noise) and 32.9\% (construction drilling noise) at -5dB. It is clear from Fig. \ref{fig:noiseTypes} that although the absolute improvement of the audio-visual model over the audio-only model is noise dependent, it generally increases as the SNR level becomes lower.   

\looseness-1
% Finally, we should also note that the audio-visual models are sometimes slightly worse than the video-only model (at -5dB). This is possibly due to the model relying internally mostly on the audio stream so when the levels of noise are high this affects the overall performance of the system.
%Ideally, an adaptive fusion model should be developed which weights each stream depending on the noise levels so the WER can never exceed the upper bound set by the video-only model.

\section{CONCLUSIONS}
In this work, we present a joint CTC/attention hybrid architecture for audio-visual speech recognition. Results on the largest publicly available database for continuous speech recognition in-the-wild (LRS2) show that the audio-visual model significantly outperforms the audio-only model especially at high levels of noise and also achieves the new state-of-the-art performance on this dataset. We use different types of noise and we show that this is true independently of the noise type considered. Finally, it would also be interesting to investigate in future work an adaptive fusion mechanism which learns to weight each modality based on the noise levels.

% Below is an example of how to insert images. Delete the ``\vspace'' line,
% uncomment the preceding line ``\centerline...'' and replace ``imageX.ps''
% with a suitable PostScript file name.
% -------------------------------------------------------------------------
% \begin{figure}[htb]

%   \centering
%   \centerline{\includegraphics[width=\linewidth]{Figures/ctc_weight}}
% %  \vspace{2.0cm}
% \caption{Word error rate as a function of the CTC weight ($\lambda$) from eq. \ref{eq:outputHypothesis}.}
% \label{fig:ctcWeight}
% %
% \end{figure}

\section{ACKNOWLEDGEMENTS}
The work of Themos Stafylakis has been funded from the European Union Horizon 2020 research and innovation programme under the Marie Sklodowska-Curie grant agreement No. 706668 (Talking Heads). The work of Pingchuan Ma has been partially funded by Honda.

% To start a new column (but not a new page) and help balance the last-page
% column length use \vfill\pagebreak.
% -------------------------------------------------------------------------
%\vfill
%\pagebreak

% References should be produced using the bibtex program from suitable
% BiBTeX files (here: strings, refs, manuals). The IEEEbib.bst bibliography
% style file from IEEE produces unsorted bibliography list.
% -------------------------------------------------------------------------
\bibliographystyle{IEEEbib}
\bibliography{mybib2}

\end{document}